\documentclass[aps,prl,twocolumn,superscriptaddress,floatfix,footinbib]{revtex4-2}

\usepackage{amsthm}
\theoremstyle{remark}
\newtheorem{theorem}{Theorem}
\newtheorem{proposition}{Proposition}
\usepackage{graphicx}
\usepackage{dcolumn}
\usepackage{bm}
\usepackage{subcaption}
\usepackage{amsmath}
\usepackage{amssymb}
\usepackage{comment}
\begin{document}

\title{\textbf{Unsupervised learning of physical concepts
via causal representation learning}}

\author{Callum Duffy}
\affiliation{Department of Physics and Astronomy, University College London, Gower Street, London WC1E 6BT, United Kingdom}
\author{Carl Fredriksson}
\affiliation{Spotify}
\author{Michael O'Riordan}
\affiliation{Spotify}
\author{Ciar\'{a}n M. Gilligan-Lee}
\affiliation{Department of Physics and Astronomy, University College London, Gower Street, London WC1E 6BT, United Kingdom}
\affiliation{Spotify}
\email{ciaranl@spotify.com}

\begin{abstract}
Identifying the latent causal variables that govern a physical system from raw observational data is a central problem across the sciences. Causal representation learning (CRL) provides a principled framework for this task: by exploiting multi-environment structure---distributions related by interventions on the underlying physical parameters---it recovers latent variables up to permutation, translation and scaling with formal identifiability guarantees. However, despite this natural alignment, no method has been developed with the explicit aim of recovering physical causal variables from data. We close this gap by introducing a contrastive CRL framework tailored to physical systems, proving identifiability of the latent causal variables under conditions directly satisfied by standard experimental designs. We validate the framework by recovering causes across a range of systems from classical and statistical mechanics to quantum field theory. These results establish CRL as a practical tool for the unsupervised discovery of physical causes.
\end{abstract}

\maketitle

The predictive power of physics rests on identifying variables whose relationships remain invariant across experimental conditions. Equations of motion and descriptions of phase transitions are useful precisely because they extend beyond the settings in which they were established. Recovering such variables directly from observations, without prior knowledge of the underlying physical description, is therefore a fundamental challenge.

This capacity for generalisation is not incidental; it is a consequence of the fact that we express physical laws in terms of causal variables, quantities whose relationships remain invariant under changes of context, measurement protocol, or boundary condition. This connection between causal structure and generalisation has a precise mathematical content within the framework of causal representation learning (CRL)~\cite{schölkopf2021causalrepresentationlearning,NIPS2016_d305281f,khemakhem2020variationalautoencodersnonlinearica,hyvarinen2019nonlinearicausingauxiliary, corcoll2026contrastive, gilligan2026unsupervised, felekis2026causal, vlontzos2026incremental, mucesh2024nature, vlontzos2023estimating, reynaud2022d, gilligan2020causing}. Despite this alignment, the field has yet to apply CRL to physical systems systematically. Prior machine learning work on physical problems has demonstrated that neural networks can detect phase transitions~\cite{Carrasquilla_2017, van_Nieuwenburg_2017,Wetzel_2017}, approximate ground states~\cite{Carleo_2017}, and extract physical parameters~\cite{PhysRevLett.124.010508}, but these approaches either rely on supervised signals or use methods such as PCA or variational autoencoders without identifiability guarantees. Separately, the CRL literature has developed extensive theory and evaluated methods on synthetic and vision-domain benchmarks~\cite{locatello2019challengingcommonassumptionsunsupervised,vonkügelgen2022selfsupervisedlearningdataaugmentations,brehmer2022weaklysupervisedcausalrepresentation}, but has not addressed the specific structure of physical systems. The question of whether the causal variables of a physical system can be provably recovered from observational data, without supervision, has remained open.

We address this challenge by introducing a contrastive CRL framework tailored to physical systems and prove formal identifiability of the latent causes. Concretely, we show that access to multiple environments constructed from physically motivated partitions of the data yields identifiability up to permutation, scaling and translation of the true latent. We additionally prove that the number of causal variables is
recoverable, providing model selection without
prior knowledge of the latent dimension. 

We validate the framework across classical mechanics, statistical mechanics, and quantum field theory: recovering spring constants and damping coefficients from trajectories, matching order parameters through phase transitions without supervision, and generalising to Schwinger model system sizes not seen during training---an empirical demonstration that causal representations generalise in a similar manner to physical laws.

\begin{figure*}[t]
  \centering
  \begin{subfigure}[b]{0.31\linewidth}
    \includegraphics[width=\linewidth]{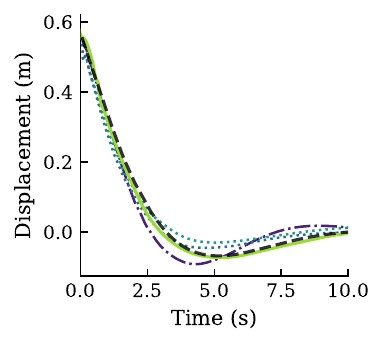}
    \caption{}\label{fig:reco_a}
  \end{subfigure}\hfill
  \begin{subfigure}[b]{0.31\linewidth}
    \includegraphics[width=\linewidth]{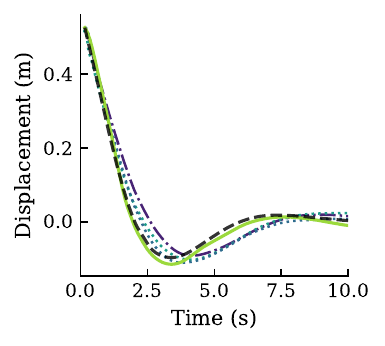}
    \caption{}\label{fig:reco_b}
  \end{subfigure}\hfill
  \begin{subfigure}[b]{0.31\linewidth}
    \includegraphics[width=\linewidth]{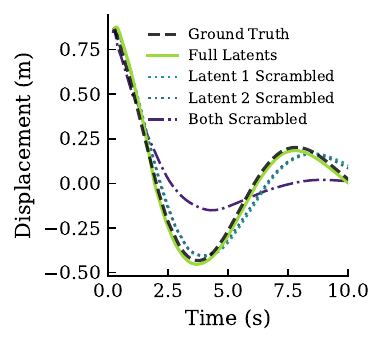}
    \caption{}\label{fig:reco_c}
  \end{subfigure}
  \caption{Damped oscillator trajectory reconstruction. Panels compare ground-truth displacement (black dashed) to reconstructions from learned concepts (solid green) and concepts scrambled by additive normal noise (dashed).}
  \label{fig:reco}
\end{figure*}

\textit{Framework---}Many quantities of physical interest, such as a spring constant or an order parameter, are not observed directly: measurements are high-dimensional nonlinear functions of them. Experiments, however, naturally provide multiple environments (temperatures, fields or control parameters) in which the mechanism is unchanged but the distribution of the latent variables shifts, and we define a physical concept as an invariant latent direction that explains these shifts.

\emph{Setup---}Let $z\in\mathbb R^n$ be the latent state, or cause, of a physical system, and let observations $x=f(z)$ be generated by a smooth injective map $f$ onto an observation manifold $\mathcal X$. We assume $f$ has a smooth inverse on its image. The $r\leq n$ physical concepts are linear projections
\begin{equation}\label{eq:concepts}
 c(x)=Af^{-1}(x)\in\mathbb R^r,\quad A\in\mathbb R^{r\times n},\quad \operatorname{rank}(A)=r.
\end{equation}

An environment does not change which physical quantities describe a system, but it can change which values of those quantities are preferentially sampled. Following the concept-conditional construction of Ref.~\cite{Rajendran2024}, imagine starting with a background ensemble $p_0(z)$ and retaining states according to noisy measurements of their concept values. If the experimental readout for concept $i$ is $a_i^\top z+\epsilon_i$, with independent Gaussian measurement noise $\epsilon_i\sim\mathcal N(0,\sigma_{ei}^2)$, conditioning on a readout $\mu_{ei}$ in environment $e$ weights the state by $\exp[-(a_i^\top z-\mu_{ei})^2/(2\sigma_{ei}^2)]$.

For $r$ such concept-selection factors, the environment-conditioned density is
\begin{equation}\label{eq:selection}
 p_e(z)\propto p_0(z)\exp\!\left[-\sum_{i=1}^r \frac{M_{ei}}{2}(a_i^\top z)^2+\sum_{i=1}^r B_{ei}(a_i^\top z)\right].
\end{equation}
The \emph{precision} $M_{ei}=\sigma_{ei}^{-2}>0$ measures how tightly environment $e$ selects concept $i$: larger $M_{ei}$ concentrates the selection around its target. The \emph{linear selection coefficient} $B_{ei}=\mu_{ei}/\sigma_{ei}^2$ records the preferred value weighted by that precision. Together they describe the width and centre of each Gaussian selection factor. Across environments the concepts $a_i^\top z$ remain the same, while the selection coefficients vary.

Choose environment $1$ as reference, and write $G_e(x)=\log[p_1(x)/p_e(x)]$ for $e=2,\ldots,m$. The common background and change-of-variables factor cancel. Introduce the $(m-1)\times r$ \emph{contrast} matrices
\begin{equation}\label{eq:contrasts}
 M_{e-1,i}:=M_{ei}^{\rm raw}-M_{1i}^{\rm raw},\qquad
 B_{e-1,i}:=B_{ei}^{\rm raw}-B_{1i}^{\rm raw}.
\end{equation}
Hereafter $M,B$ denote these contrasts. The observable vector of log-density ratios has the form
\begin{equation}\label{eq:phi}
\begin{aligned}
 \Phi(x)&:=(G_2(x),\ldots,G_m(x))^\top
 =F(c(x)), \\
 F(u)&=\frac12 M u^{\odot2}-Bu+\alpha,
 \end{aligned}
\end{equation}
where $u^{\odot2}$ denotes componentwise squaring and $\alpha$ contains normalisation constants.

Ref.~\cite{Rajendran2024} proved that underlying concepts can be uniquely recovered from observations across sufficiently diverse environments, up to the unavoidable ambiguities of relabelling, rescaling, and shifting the concepts. However, their result requires a pairwise-separation assumption, which requires that for all concept pairs, there exists at least one environment that selects one but not the other. Such selective absence is unnatural when the same physical quantities govern a system under every environment. Here we prove a theorem that establishes the same conclusion, but removes the pairwise-separation assumption, making our result applicable to physical systems. Our proof is given in the Appendix.

\begin{theorem}[Observational identification of concepts]\label{thm:atomic}
Suppose $m=r+2$ and the model \eqref{eq:concepts}--\eqref{eq:phi} holds on $\mathbb R^n$, with smooth full-rank concept map $c=Af^{-1}$ and contrast matrices $M,B\in\mathbb R^{(r+1)\times r}$. Assume
\begin{equation}\label{eq:rankkernel}
 \operatorname{rank}(M)=r,\text{ and } \exists\,v\in\ker(M^\top)\setminus\{0\}: (B^\top v)_i\neq0 %\quad(i=1,\ldots,r).
\end{equation}
If a second model of the same form, with $r$ concepts, a smooth concept map $\widetilde c$ taking every value in $\mathbb R^r$, and contrast matrices satisfying \eqref{eq:rankkernel}, induces the same observable environment distributions, then
\begin{equation}\label{eq:ambiguity}
 \widetilde c(x)=PDc(x)+t %\qquad (x\in\mathcal X),
\end{equation}
with $P$ is a permutation matrix, $D$ an invertible diagonal matrix, and $t\in\mathbb R^r$. Hence concepts are observationally identified up to permutation, scaling and translation.  %In particular, the number of required environments depends on the number $r$ of concepts, not on the latent dimension $n$.
\end{theorem}

The assumptions express complementary forms of \emph{sufficient diversity}. Full rank of $M$ requires the environments to change the precision of the $r$ concepts in linearly independent ways. Physically, the kernel-vector condition says preferred concept values must contain variation that cannot be explained solely by changes in how tightly the environments constrain them. One can show that the full rank condition implies the kernel-vector condition holds except on a set of measure zero, by a genericity argument analogous to that of Ref.~\cite{Rajendran2024}.

\textit{Contrastive estimation---}We operationalise Theorem 1 by approximating each $G_e(x)$ with a parametric classifier mirroring the quadratic form of the true log-density ratio,
\begin{equation}\label{eq:classifier}
  g_e(x;\theta) = \alpha^e
  - \textstyle\sum_j \bigl(\beta_j^e\, h_j^\theta(x)\bigr)^2
  + \textstyle\sum_j \gamma_j^e\, h_j^\theta(x),
\end{equation}
where $h^\theta(x)$ is a neural encoder and $\{\alpha^e,\beta_j^e,\gamma_j^e\}$ are learnable environment-specific parameters. Encoder and classifier are trained jointly by minimising a contrastive binary cross-entropy loss
\begin{equation}\label{eq:loss}
  \mathcal{L} = \sum_e -\mathbb{E}_{j\sim U(\{1,e\})}
  \mathbb{E}_{x\sim p^j}\!\left[\log
  \frac{e^{\mathbf{1}_{j=e} g_e(x)}}{1+e^{g_e(x)}}\right].
\end{equation}
At optimality $g_e$ recovers $\log [p_e(x)/p_1(x)]=-G_e(x)$.

\textit{Determining the number of concepts---}The number of concepts that the experimental design can recover is itself usually unknown. We show the dimension $r$ of the environment-resolved concept space can be read off from the rank of a density-ratio Jacobian. Differentiating Eq.~\eqref{eq:phi} with respect to observations gives
\begin{equation}\label{eq:jacfactor}
 D\Phi(x)=K(c(x))\,Dc(x),\quad K(u)=M\operatorname{diag}(u)-B.
\end{equation}

\begin{proposition}[Counting concepts via density-ratio Jacobians]\label{prop:count}
Suppose $c:\mathcal X\to\mathbb R^r$ is smooth, takes every value in $\mathbb R^r$, has Jacobian rank $r$ everywhere, and the ratios have form \eqref{eq:phi}. If $\operatorname{rank}(M)=r$ and $m-1\ge r$, then
\begin{equation}\label{eq:rankresult}
 \operatorname{rank}D\Phi(x)=r
\end{equation}
for all $x$ outside the inverse image under $c$ of a proper algebraic subset of $\mathbb R^r$.
\end{proposition}

\begin{figure*}[t]
  \centering
  \begin{subfigure}[b]{0.45\linewidth}
    \includegraphics[width=\linewidth]{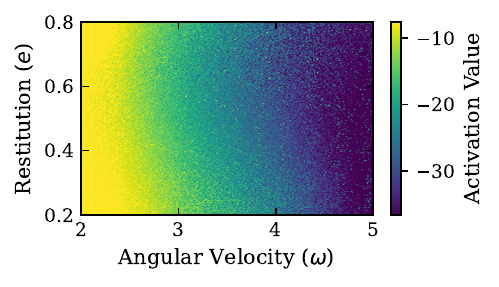}
    \caption{Concept $h_1$ over $(\omega,e)$.}
  \end{subfigure}\hfill
  \begin{subfigure}[b]{0.45\linewidth}
    \includegraphics[width=\linewidth]{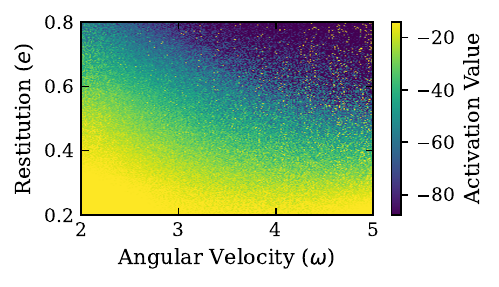}
    \caption{Concept $h_2$ over $(\omega,e)$.}
  \end{subfigure}
  \caption{Latent activations for the rod--ball collision over a grid of restitution coefficient $e$ and pre-collision angular velocity $\omega$. Panels (a) and (b) show the learned concepts $h_1$ and $h_2$, respectively.}
  \label{fig:rodbball}
\end{figure*}

\textit{Experiments---}We quantify representation quality with the mean correlation coefficient (MCC)~\cite{hyvarinen2019nonlinearicausingauxiliary}, invariant to permutation and scaling, which will indicate whether our models recover the true underlying physics of the system when compared to the ground truth. Architectures and training details can be found in the Appendix.

\textit{Damped harmonic oscillator---}The damped oscillator governed by the equation of motion $m\ddot x + b\dot x + \kappa x = 0$ is a minimal testbed: the concepts are the spring constant and damping coefficient $(\kappa,b)$, and $f$ maps them to the observed trajectory $x(t)$ via the analytic solution. Three environments with nominal $(\kappa,b)\in\{(1.0,1.0),(3.5,1.0),(1.0,3.5)\}$ and Gaussian perturbations, together with the base distribution, give $m=4=r+2$ environments with $\mathrm{rank}(M)=r=2$, as Theorem~1 requires. An MLP encoder with two-dimensional output is trained contrastively on 5{,}000 trajectories per environment, where the amplitude and phase are also randomly sampled along with $\kappa$ and $b$.

Against the ground-truth $(\kappa,b)$ the model attains an MCC of $0.895\pm0.019$ on the environments and $0.708\pm0.027$ on the base distribution (Table~\ref{tab:mcc_all}), consistent with prior identifiability results in CRL~\cite{Rajendran2024}. To confirm the concepts carry dynamics-relevant information, we train a downstream autoregressive Transformer to reconstruct held-out trajectory segments from the learned concepts and the initial amplitude alone (Fig.~\ref{fig:reco}). Scrambling either concept individually, by adding standard normal noise to it, raises the reconstruction MSE by more than an order of magnitude (from $6.72\times10^{-4}$ to $1.87\times10^{-2}$), and scrambling both raises it to $3.90\times10^{-2}$; the symmetry between the single-scramble conditions confirms the dimensions encode distinct physical information.

Overcomplete encoders trained on the oscillator yield exactly two dominant singular values of $D\Phi$ for every width, as Proposition~\ref{prop:count} predicts (see Appendix).

\textit{Rotating rod--ball collision---}We next study a planar collision between a freely rotating rod (length $L=1$\,m, mass $m_{rod}=1$ kg) and an impacting point particle ($m_{ball}=1$ kg), where the causal concepts are the conserved angular momentum $J$ and the dissipative coefficient of restitution $e$. These play distinct physical roles: angular momentum about the rod's pivot is strictly conserved ($I\omega^- + m L v_t^- = I\omega^+ + m L v_t^+$), while the energy lost to inelasticity is governed by $e = -v_{\mathrm{rel}}^+ / v_{\mathrm{rel}}^- \in [0,1]$. Three environments with nominal $(\omega, e)\in\{(2.0,0.8),(5.0,0.2),(2.0,0.2)\}$ and Gaussian perturbations are constructed, %($\sigma_\omega=0.1$, $\sigma_e=0.01$) 
which, together with the base distribution, give \(m=4=r+2\) environments. The initial positions of both bodies chosen at random. From a sequence of 50 pre- and post-collision observations $o_i = [t_i,\; \mathbf{x}_{\mathrm{ball},i},\; \mathbf{x}_{\mathrm{rod},i},\; \omega_i]$ over 3 seconds, the model must infer both concepts. The MLP achieves an MCC of $0.810\pm0.137$ for the environments and $0.746\pm0.149$ for the base distribution (Table~\ref{tab:mcc_all}). Latent activation heatmaps over the $(\omega, e)$ grid (Fig.~\ref{fig:rodbball}) show that the gradients of the two learned concepts are approximately orthogonal, with one tracking angular momentum (controlled by the pre-collision angular velocity $\omega$) and the other tracking $e$, confirming that the framework disentangles the conserved quantity from the dissipative parameter as predicted by Theorem~1.
\begin{figure*}[t]
  \centering
  \begin{subfigure}[b]{0.32\linewidth}
    \includegraphics[width=\linewidth]{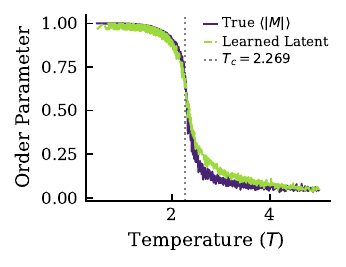}
    \caption{Ferromagnetic Ising.}\label{fig:ferro}
  \end{subfigure}\hfill
  \begin{subfigure}[b]{0.32\linewidth}
    \includegraphics[width=\linewidth]{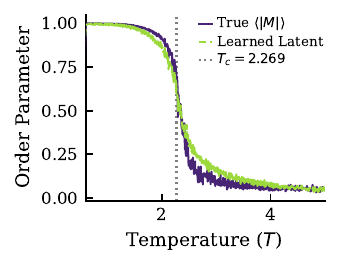}
    \caption{Antiferromagnetic Ising.}\label{fig:antiferro}
  \end{subfigure}\hfill
  \begin{subfigure}[b]{0.32\linewidth}
    \includegraphics[width=\linewidth]{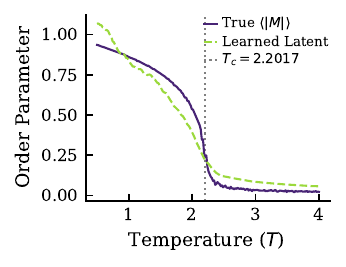}
    \caption{3D XY model.}\label{fig:xy}
  \end{subfigure}
  \caption{Phase transitions in statistical mechanics models: (a) 2D Ising ferromagnetic, (b) 2D Ising antiferromagnetic, and (c) 3D XY model. In each panel, the ground-truth order parameter (solid purple) and the learned concept (dashed green) are plotted against temperature. Vertical dashed line is critical temperature $T_c$.}
  \label{fig:stat_mech}
\end{figure*}

\textit{Statistical mechanics}---We study the two-dimensional Ising and three-dimensional XY models, where $f$ maps a single scalar concept (the order parameter) to high-dimensional spin configurations. We set $k_B=1$ and simulate $75{,}000$ samples for each system via Monte Carlo~\cite{PhysRevB.94.165134, Metropolis01091949}, split evenly between training and testing. 

For the Ising model ($s_i \in \{\pm 1\}$), the Hamiltonian is $H=-J\sum_{\langle ij\rangle}s_is_j$, with nearest-neighbour interactions. For $J>0$ the system is ferromagnetic (favouring aligned spins), and for $J<0$ it is antiferromagnetic (favouring alternating spins). Below $T_c \approx 2.269$, the global $\mathbb{Z}_2$ symmetry ($s_i \to -s_i$) is spontaneously broken~\cite{PhysRev.65.117}. The target concepts are the thermal order parameter $\langle |M| \rangle$ with $M = \frac{1}{N}\sum s_i$ for $J>0$, and the staggered magnetisation $M_{\mathrm{st}} = M(S \odot A)$ with $A_{ij} = (-1)^{i+j}$ for $J<0$. We process the $28 \times 28$ configurations ($784$ binary variables) with a 2D CNN.

The 3D XY model extends this to continuous $U(1)$ symmetry. Spins are planar unit vectors $\mathbf{s}_i = (\cos\theta_i, \sin\theta_i)$ with $\|\mathbf{s}_i\|=1$, and the Hamiltonian is $H=-\sum_{\langle ij\rangle}\mathbf{s}_i \cdot \mathbf{s}_j$ ($J=1$), with nearest-neighbour interactions. The $U(1)$ symmetry breaks below $T_c \approx 2.20$~\cite{GOTTLOB1993593}, with the continuous order parameter given by $M = \langle\| \frac{1}{N} \sum \mathbf{s}_i \|\rangle$. We simulate on an $L=16$ cubic lattice with periodic boundary conditions, generating 3D grids of 2D spin vectors ($16^3 \times 2 = 8192$ continuous variables), processed with a 3D CNN.

For all systems, two environments are constructed by sampling temperatures from $\mathcal{N}(1.5,0.4^2)$ (ordered) and $\mathcal{N}(3.5,0.4^2)$ (disordered); with the base distribution this gives $m=3=r+2$ and $\mathrm{rank}(M)=1$. Across all experiments, the 1D latent output faithfully tracks the ground-truth order parameter through the phase transition despite never seeing it during training (shown in Fig.~\ref{fig:stat_mech}). By rescaling the latent variable using a single reference temperature ($T=0.5$), we obtain a curve that closely matches the true $M(T)$, demonstrating that the framework successfully reconstructs the order parameter up to an affine transformation. MCCs exceed $0.93$ for all three systems (Table~\ref{tab:mcc_all}).

\begin{figure}[b]
  \centering
  \includegraphics[width=0.9\linewidth]{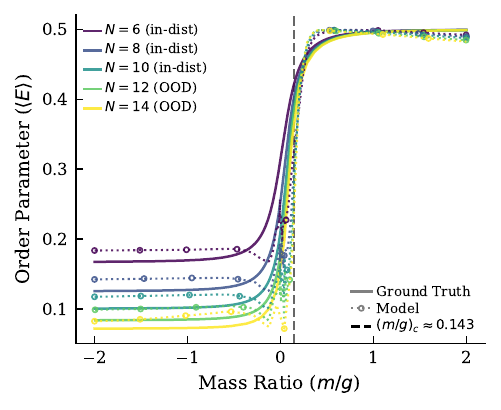}
  \caption{Phase transition in the Schwinger model. The ground-truth order parameter $\langle E \rangle$ (solid lines) is compared against the learned concepts (dotted lines). System sizes $N \in \{6,8,10\}$ (in-distribution) and $N \in \{12,14\}$ (out-of-distribution) are distinguished by colour. Vertical dashed line is critical mass ratio $(m/g)_c$. % \approx 0.143$.
  }
  \label{fig:schwinger}
\end{figure}

\textit{Schwinger model}---Finally, we study the (1+1)-dimensional Schwinger model, a lattice gauge theory with fermionic degrees of freedom. Machine learning is widely applied to quantum systems~\cite{Carleo_2019, dunjko2017machinelearningartificial, Bennewitz:2021jqi}, but without identifiability guarantees. Here we apply CRL to a quantum system for the first time, providing a principled and unsupervised route to recovering the causal concept governing its phase structure directly from observables.

Instead of the full $2^N$-dimensional quantum ground state, the framework observes only a nonlinear projection $\pi$: local Pauli expectation values $(\langle Z\rangle, \langle X\rangle, \langle Y\rangle)$ as node features, and nearest-neighbour correlators $(\langle XX\rangle, \langle YY\rangle)$ as edge features. A lattice Hamiltonian for the continuum Schwinger model can be obtained via the staggered formalism~\cite{PhysRevD.11.395}. In $1$D, with open boundary conditions and Gauss's law, all gauge fields can be integrated out. Following a Jordan--Wigner transformation~\cite{Jordan:1928wi}, the lattice Hamiltonian is 
\begin{equation}
\begin{split}
  H &= J\!\sum_{n=0}^{N-2}\!\left(\sum_{i=0}^n\frac{Z_i+(-1)^i}{2}
  +\frac{\theta}{2\pi}\right)^{\!2}
  +\frac{m}{2}\sum_{n=0}^{N-1}(-1)^nZ_n\\
  &+\frac{w}{2}\!\sum_{n=0}^{N-2}\!(X_nX_{n+1}+Y_nY_{n+1}),
\end{split}
\end{equation}
with $J=ag^2/2$ and $w=1/(2a)$. The target concept is the average electric field $E=N^{-1}\sum_n\langle L(n)+\alpha\rangle$. Using Gauss's law ($L(n)+\alpha = \sum_{i=0}^n [Z_i+(-1)^i]/2$), $E$ is computable entirely from the local $Z$-expectation node features. At topological angle $\theta=\pi$ ($\alpha=1/2$), the system undergoes a quantum phase transition at $(m/g)_c\approx0.143$ (for a finite lattice $N=8$, $ag=2$)~\cite{PhysRevResearch.5.043250}, separating a symmetric phase ($\langle E\rangle=0$) from a symmetry-broken one.

We compute exact ground states for $N \in \{6,8,10,12,14\}$, generating 10,000 samples per environment by drawing mass ratios from both phases. A graph neural network (GNN) with edge-aware message passing and global mean pooling maps the variable-sized graphs to a size-invariant 1D latent.

Crucially, we train the network exclusively on $N \in \{6,8,10\}$; the larger sizes $N \in \{12,14\}$ are held out entirely to test out-of-distribution (OOD) generalisation. The learned latent faithfully captures the transition for training sizes and continues to track $\langle E\rangle$ for the unseen lattices (Fig.~\ref{fig:schwinger}). The finite-size artefact, a non-zero latent in the symmetric phase, decreases monotonically with $N$ even for the OOD systems, exactly as expected from finite-size scaling. This confirms the representation encodes intrinsic phase structure rather than memorising size-specific patterns. OOD performance is comparable to in-distribution performance (Table~\ref{tab:mcc_all}). By identifying quantum phase structure from local observables, this experiment demonstrates that causal representations generalise in the same way physical laws do. This aligns with a broader programme of learning from quantum-generated data: recent hardware experiments have likewise trained classical models on local observables of many-body ground states and found accurate prediction at unseen Hamiltonian parameters, including out of distribution~\cite{Jaderberg:2026ewk}, motivating a path toward training regimes beyond the reach of classical approximation.

\begin{table}[t]
\caption{\label{tab:mcc_all}
MCC between learned concepts and ground-truth on held-out test data. One model is trained on all environments; ``Env.'' is evaluated on samples from the selection environments, ``Base'' on samples from the base distribution. In-distribution (ID) and out-of-distribution (OOD) sizes are noted.}
\begin{ruledtabular}
\begin{tabular}{llcc}
System & Enc. & Env. & Base \\
\colrule
Oscillator & MLP & $0.895\pm0.019$ & $0.708\pm0.027$ \\
Rod--ball & MLP & $0.810\pm0.137$ & $0.746\pm0.149$ \\
Ferro.\ Ising & CNN & --- & $0.985\pm0.005$ \\
AF Ising & CNN & --- & $0.988\pm0.003$ \\
3D XY & 3D CNN & --- & $0.935\pm0.042$ \\
\colrule
\multicolumn{4}{l}{\textit{Schwinger model}} \\
$N=6$ (ID) & GNN & --- & $0.879\pm0.137$ \\
$N=8$ (ID) & GNN & --- & $0.910\pm0.081$ \\
$N=10$ (ID) & GNN & --- & $0.924\pm0.071$ \\
$N=12$ (OOD) & GNN & --- & $0.917\pm0.092$ \\
$N=14$ (OOD) & GNN &  --- & $0.923\pm0.086$ \\
\end{tabular}
\end{ruledtabular}
\end{table}
\textit{Conclusion---}We have shown that causal representation learning recovers physically meaningful concepts across classical mechanics, statistical mechanics, and quantum field theory, with identifiability guarantees. The framework requires no supervision, handles high-dimensional and variable-sized systems, and provides a way to determine the number of latent concepts.

Beyond recovering known physics, a central motivation for this framework is the discovery of new physical concepts in systems where theory is absent or incomplete. Given multi-environment measurements of a system whose governing equations are unknown---a condensed-matter system in phases whose microscopic origin is not yet understood, or astronomical observations across distinct dynamical regimes---the framework extracts the causal variables whose values distinguish the environments. These constitute hypotheses around which a future theory might be organised, and because they are causal they should generalise to new conditions in the same way physical laws do. 

\begin{acknowledgments}
\textit{Acknowledgments---}C.D. acknowledges support from the CDT-DIS.
\end{acknowledgments}

\bibliography{apssamp}

\onecolumngrid
\vspace{2em}

\begin{center}
    \textbf{\large Appendix}
\end{center}

\vspace{1em}

\twocolumngrid
\appendix

\section*{Proof of Theorem~\ref{thm:atomic}}
\noindent\textit{Proof strategy.} Consider weighted sums of the observable environment log-density ratios. Each concept contributes a quadratic term, whose coefficient depends on the weights, and a linear term. Full rank of the quadratic contrast matrix lets us choose weights that cancel the quadratic term of any selected concept without cancelling those of the others. The kernel-vector condition supplies a direction in weight space that leaves every quadratic coefficient unchanged but changes the selected linear coefficient. We can therefore arrange either that the selected concept enters only linearly, in which case the weighted ratio has no stationary point, or that both its terms vanish, in which case stationary points are possible again. The weights at which these changes occur are observable properties of the density ratios.

For weights with nonzero quadratic coefficients for every concept, the stationary equations can be solved explicitly: each stationary concept value is the ratio of its linear and quadratic coefficients. The two representations describe exactly the same function of the observations, so the points at which its derivative vanishes are the same, regardless of how those observations are parametrised. Comparing the two expressions for the concept values at these common stationary observations, using the coefficient relations just established, gives a coordinate-wise affine transformation (up to permutation). Finally, varying the weights produces stationary concept values on a dense subset of the concept space; continuity extends that affine relation to every observation. This last step turns identification at specific stationary observations into a full identifiability result.

\begin{proof}
We compare two observationally equivalent concept factorisations $\Phi=F\circ c=\widetilde F\circ\widetilde c$. Write $m_i,b_i\in\mathbb R^{r+1}$ for the columns of $M,B$. Because $c(x)=Af^{-1}(x)$ and $A$ has rank $r$, its derivative $Dc(x)$ has rank $r$ everywhere. For any smooth $H:\mathbb R^r\to\mathbb R$, the chain rule gives $d(H\circ c)=(Dc)^\top dH$. Since $(Dc)^\top$ is injective, an observation is a stationary point of $H\circ c$ exactly when its concept value is a stationary point of $H$. The stationary observations therefore form sets of the form $\{x:c(x)=u\}$, where $u$ is a stationary concept value. This lets us reason using only $r$ concept coordinates even when $r<n$.

For $w\in\mathbb R^{r+1}$ define $g_w(x)=w^\top\Phi(x)$. In concept coordinates,
\begin{equation}\label{eq:gw}
 g_w=\frac12\sum_{i=1}^r(w^\top m_i)c_i^2-\sum_{i=1}^r(w^\top b_i)c_i+\mathrm{const},
\end{equation}
so its stationary concept values solve
\begin{equation}\label{eq:stationary}
 (w^\top m_i)c_i=w^\top b_i\qquad (i=1,\ldots,r).
\end{equation}

\emph{Step 1: Identify the quadratic contrast directions.} Put $H_i=\ker(m_i^\top)\subset\mathbb R^{r+1}$. Full column rank of $M$ makes the $H_i$ distinct. By the kernel-vector condition, $b_i\notin\operatorname{span}(m_i)$: otherwise $v^\top b_i=0$. Outside $\bigcup_iH_i$, Eq.~\eqref{eq:stationary} has exactly one solution, hence $g_w$ has a nonempty set of stationary observations. On $H_i$, away from the other $H_j$, Eq.~\eqref{eq:stationary} has no solution whenever $w^\top b_i\neq0$; these weights form a relatively open dense subset of $H_i$. Hence the closure of the observable set of weights with no stationary point is precisely $\bigcup_iH_i$. Its distinct linear hyperplane components are uniquely determined, so comparison with the second representation gives, after permuting its coordinates,
\begin{equation}\label{eq:mcompare}
 \widetilde m_i=\lambda_i m_i,\qquad\lambda_i\neq0.
\end{equation}

\emph{Step 2: Identify linear contrast directions modulo the quadratic directions.} For $w\in H_i\setminus\bigcup_{j\neq i}H_j$, Eq.~\eqref{eq:stationary} has a solution exactly when $w^\top b_i=0$. This observable condition determines the hyperplane $H_i\cap\ker(b_i^\top)$ inside $H_i$ (also for $r=1$, when $H_i$ is one-dimensional and this intersection is $\{0\}$). The restrictions of $b_i$ and $\widetilde b_i$ to $H_i$ therefore have the same kernel and are nonzero. Elementary linear algebra yields
\begin{equation}\label{eq:bcompare}
 \widetilde b_i=\mu_i b_i+\nu_i m_i,\qquad\mu_i\neq0.
\end{equation}

\emph{Step 3: Identify the coordinate relation on the observable set of stationary observations.} For $w\notin\bigcup_iH_i$, the stationary concept value is unique:
\begin{equation}\label{eq:critcoord}
 u_i(w)=\frac{w^\top b_i}{w^\top m_i}.
\end{equation}
The two models have the same set of stationary observations $\{x:c(x)=u(w)\}=\{x:\widetilde c(x)=\widetilde u(w)\}$. From Eqs.~\eqref{eq:mcompare}--\eqref{eq:critcoord}, for every observation $x$ in this set,
\begin{equation}\label{eq:affineonstationary}
 \widetilde c_i(x)=\frac{\mu_i}{\lambda_i}c_i(x)+\frac{\nu_i}{\lambda_i}.
\end{equation}

\emph{Step 4: Show these sets of stationary observations are dense in concept space.} Since $\operatorname{rank}M=r$, its left kernel is spanned by $v$. Write
\begin{equation}\label{eq:decomp}
 B=MR+vq^\top,\qquad q_i=\frac{v^\top b_i}{v^\top v}\neq0
\end{equation}
for some $R\in\mathbb R^{r\times r}$. This decomposition exists because the columns of $[M\ v]$ form a basis of $\mathbb R^{r+1}$. Let $L(u)=\operatorname{diag}(u)-R$. For any $u$ with $\det L(u)\neq0$, choose $w$ uniquely by imposing
\begin{equation}\label{eq:choosew}
 w^\top v=1,\qquad w^\top M=q^\top L(u)^{-1}.
\end{equation}
These conditions are consistent because $[M\ v]$ is invertible. Substitution into $w^\top[M\operatorname{diag}(u)-B]=0$ shows that $u$ is a stationary concept value. It is the \emph{unique} stationary concept value if every entry of $w^\top M$ is nonzero. Using the adjugate formula, this holds whenever
\begin{equation}\label{eq:denseU}
 \det L(u)\neq0,\qquad P_i(u):=[q^\top\operatorname{adj}L(u)]_i\neq0\quad\forall i.
\end{equation}
Each expression in Eq.~\eqref{eq:denseU} is a polynomial and none is identically zero: along $u=t\bm1$, $\det(tI-R)=t^r+O(t^{r-1})$ and $P_i(t\bm1)=q_i t^{r-1}+O(t^{r-2})$ as $t\to\infty$ (for $r=1$, $P_1=q_1$). Thus the set $U$ defined by Eq.~\eqref{eq:denseU} is open dense in $\mathbb R^r$. Every $u\in U$ is the unique stationary concept value for some $w\notin\bigcup_iH_i$. Eq.~\eqref{eq:affineonstationary} therefore holds on $c^{-1}(U)$.

\emph{Step 5: Extend to every observation.} Since $Dc$ has rank $r$ everywhere, the image under $c$ of any nonempty open neighbourhood in $\mathcal X$ contains an open subset of $\mathbb R^r$ (by the constant-rank theorem). That subset intersects the dense set $U$, so $c^{-1}(U)$ is dense in $\mathcal X$. Both $c$ and $\widetilde c$ are continuous. Eq.~\eqref{eq:affineonstationary} consequently extends from $c^{-1}(U)$ to all of $\mathcal X$. Restoring the coordinate permutation gives Eq.~\eqref{eq:ambiguity}. No global inverse for the concept map, or additional rank condition on $DF$, has been used.
\end{proof}

\section*{Proof of Proposition 1}
\begin{proof}
The chain rule gives Eq.~\eqref{eq:jacfactor}. Because $Dc(x)$ has full row rank, $\operatorname{rank}D\Phi(x)=\operatorname{rank}K(c(x))$. Since $M$ has full column rank, some $r\times r$ row minor $M_I$ is nonsingular. The corresponding minor of $K(u)$ is
\begin{equation}\label{eq:minor}
 p(u)=\det\bigl(M_I\operatorname{diag}(u)-B_I\bigr).
\end{equation}
The coefficient of the monomial $u_1\cdots u_r$ in $p$ equals $\det M_I\neq0$. Thus $p$ is a nonzero polynomial. Outside its proper algebraic zero set, $K(u)$ has rank $r$, and therefore so does $D\Phi(x)$. Because $Dc$ has full row rank everywhere, local coordinates can be chosen in which the first $r$ coordinates are $c_1,\ldots,c_r$. In such coordinates the inverse image of a proper algebraic subset of $\mathbb R^r$ has empty interior and measure zero (by Fubini's theorem).

\end{proof}

\section*{Recovering the number of concepts}

\begin{figure}[t]
  %\centering
  \includegraphics[width=\linewidth]{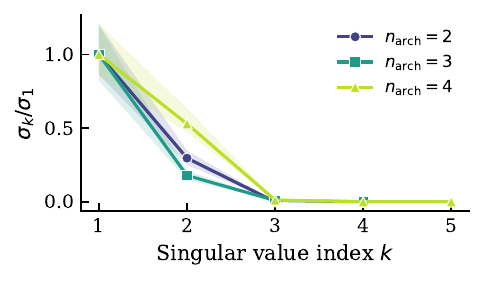}
  \caption{Jacobian singular value spectrum across 300 held-out points for models trained with $n_{\mathrm{arch}}\in\{2,3,4\}$ ($m=n_{\mathrm{arch}}+1$ environments). Solid lines show the mean, shaded regions its standard deviation.}
  \label{fig:svd}
\end{figure}

To validate Proposition~1 we train overcomplete models with encoder widths $n_{\mathrm{arch}}\in\{2,3,4\}$ and $m=n_{\mathrm{arch}}+1$ environments, computing the singular value spectrum of $D\Phi$ at 300 held-out points (Fig.~\ref{fig:svd}). The damped oscillator activates two concept directions. For every encoder width, a clear spectral gap isolates exactly two dominant singular values, with the surplus dimensions collapsing rather than absorbing spurious structure. The spectrum therefore identifies the number of recoverable concepts independently of the architectural choice. The leading singular value markedly exceeds the second, reflecting that the trajectory is more sensitive to the spring constant (which sets the oscillation frequency) than to the damping coefficient (which sets the slow decay envelope).

\section*{Experimental details}\label{appx:exp_details}
\begin{table*}[t!]
\caption{\label{tab:experimental_details} Summary of experimental configurations, including dataset generation, architectures, and hyperparameters.
}
\begin{ruledtabular}
\begin{tabular}{p{0.16\linewidth} p{0.18\linewidth} p{0.20\linewidth} p{0.20\linewidth} p{0.16\linewidth}}
Parameter & Damped Oscillator & Rod--Ball Collision & Statistical Mechanics & Schwinger Model \\
\colrule
\multicolumn{5}{l}{\textbf{I. Dataset Generation and Physical Parameters}} \\
\colrule
\textbf{Physical Constants} & Mass $m=4$ & $m_{\mathrm{rod}}=1$, $m_{\mathrm{ball}}=1$,\newline $L=1$ & $J=\pm 1$ (Ising),\newline $J=1$ (XY) & $g=1.0$, $\theta=\pi$,\newline $(m/g)_c=0.143$ \\
\textbf{Environments ($m$)} & 4 (3 + base) & 4 (3 + base) & 3 (2 + base) & 3 (2 + base) \\
\textbf{Samples per Env} & $5{,}000$ & $10{,}000$ & $25{,}000$ & $10{,}000$ \\
\textbf{Train/Val/Test} & 50\% / 25\% / 25\% & 50\% / 25\% / 25\% & 50\% / 0\% / 50\% & 50\% / 0\% / 50\% \\
\textbf{Observation Grid} & 50 pts over $[0, 10]$\,s & 50 pts over $[0, 3]$\,s & MCMC: 5,000 therm.,\newline 200 skip & ED for\newline $N \in \{6, 8, 10\}$ \\
\textbf{Nominal Concepts} & $(\kappa, b) \in \{(1, 1),$\newline $(3.5, 1), (1, 3.5)\}$ & $(\omega, e) \in \{(2.0, 0.8),$\newline $(5.0, 0.2), (2.0, 0.2)\}$ & $T \sim \mathcal{N}(1.5, 0.4^2),$\newline $\mathcal{N}(3.5, 0.4^2)$ & $m/g \sim \mathcal{N}(-1.0, 0.4^2),$\newline $\mathcal{N}(1.0, 0.4^2)$ \\
\textbf{Concept Noise} & $\sigma_\kappa=0.1$, $\sigma_b=0.2$ & $\sigma_\omega=0.1$, $\sigma_e=0.01$ & N/A (distribution) & N/A (uniform) \\
\textbf{Initial Conditions} & $A \sim U(0.5, 1.5)$,\newline $\phi \sim U(0, 3)$ & $v_b=[0.5, 0.8]$,\newline random pos. & Random spin init. & Quantum ground state \\
\colrule
\multicolumn{5}{l}{\textbf{II. Neural Network Architectures}} \\
\colrule
\textbf{Encoder Type} & MLP & MLP & 2D CNN / 3D CNN & GNN (GATv2Conv) \\
\textbf{Hidden Layers} & 3 & 3 & 3 Conv + FC bottleneck & 3 message-passing \\
\textbf{Hidden Dims} & 64 & 64 & Channels: $\{16, 32, 64\}$\newline / $\{8, 16, 32\}$ & 64, 4 attention heads \\
\textbf{Activation} & LeakyReLU(0.2) & LeakyReLU(0.2) & LeakyReLU(0.2) & ReLU \\
\textbf{Pooling Type} & N/A & N/A & AdaptiveAvgPool & Global mean pool \\
\textbf{Latent Size ($n_{\mathrm{arch}}$)} & 2 & 2 & 1 & 1 \\
\textbf{Auxiliary Model} & Traj. Regressor & None & None & None \\
\textbf{Aux. Specs} & 1-layer Transformer\newline ($d_{\mathrm{model}}=64$) & --- & --- & --- \\
\colrule
\multicolumn{5}{l}{\textbf{III. Optimisation and Training Hyperparameters}} \\
\colrule
\textbf{Optimiser} & Adam & Adam & Adam & Adam \\
\textbf{Encoder LR} & $10^{-4}$ & $10^{-4}$ & $10^{-4}$ & $10^{-4}$ \\
\textbf{Classifier LR} & $5 \times 10^{-2}$ & $5 \times 10^{-2}$ & $10^{-3}$ & $10^{-3}$ \\
\textbf{Weight Decay} & $10^{-5}$ & $10^{-4}$ & $10^{-4}$ & $10^{-4}$ \\
\textbf{Batch Size} & 500 & 256 & 64 & 128 \\
\textbf{Epochs} & 100 & 170 & 5 (Ising), 10 (XY) & 52 \\
\textbf{LR Scheduler} & None & Cosine Annealing\newline ($T_{\max}=50$) & Cosine Annealing\newline ($T_{\max}=100$) & Cosine Annealing\newline ($T_{\max}=100$) \\
\end{tabular}
\end{ruledtabular}
\end{table*}

To ensure reproducibility, Table~\ref{tab:experimental_details} details the complete experimental configurations for the physical systems studied in the main text: the damped harmonic oscillator, the inelastic rod--ball collision, the classical statistical mechanics models, and the quantum Schwinger model. The table summarises the physical dataset generation parameters, neural network architectures, %(including all concept encoders and auxiliary probes),  
and hyperparameters used during training.

\end{document}